\documentclass{article}
\usepackage{fullpage}
\usepackage{prooftree}
\usepackage{fancybox}\fboxrule=2\fboxrule
\usepackage[latin1]{inputenc}
\usepackage{amssymb,dsfont}
\usepackage[all]{xy}\SelectTips{cm}{} \CompileMatrices
\usepackage{qsymbols}
\usepackage{pdflscape}
\usepackage{graphics}
\usepackage{alltt}
\usepackage{url}

\newcommand{\Coq}{\textsc{Coq}}
\newcommand{\el}{epistemic logic}
\newcommand{\elck}{common knowledge logic}
\newcommand{\Elck}{Common knowledge logic }
\newcommand{\pourtout}{\texttt{$\backslash$-/}}
\newcommand{\MvdH}{$\mathbb{TEC}_G$}
\newcommand{\MvdHalt}{$\mathbb{TEC}_G'$}
\newcommand{\ComK}{$\mathbb{CK}_G$}
\newcommand{\four}{\ensuremath{\mathbf{4}}}

\newcommand{\non}{\begin{math}\neg\end{math}}
\newcommand{\axvsck}{\emph{external vs internal}}
\newcommand{\FP}{\textbf{FB}}
\newcommand{\LFP}{\textbf{LFB}}
\newcommand{\MP}{\textbf{MP}}
\newcommand{\KG}{\textbf{KG}}

\begin{document}

\title{\textsf{\textbf{\large LIP Research Report RR2007-51}}\\[5pt]
Common knowledge logic
  \\in a higher order proof assistant? %
}

\author{Pierre Lescanne
%
  Laboratoire de l'Informatique du Parallélisme,\\
  École Normale Supérieure de Lyon \\
  46, Allée d'Italie, 69364 Lyon 07, FRANCE\\[2pt]
\texttt{Pierre.Lescanne@ens-lyon.fr}
}

\maketitle

\bigskip

\begin{abstract}
  This paper presents experiments on common knowledge logic, conducted
  with the help of the proof assistant \Coq.  The main feature of
  common knowledge logic is the eponymous modality that says that a
  group of agents shares a knowledge about a certain proposition in a
  inductive way.  This modality is specified by using a fixpoint
  approach.  Furthermore, from these experiments, we discuss and
  compare the structure of theorems that can be proved in specific
  theories that use common knowledge logic.  Those structures
  manifests the interplay between the theory (as implemented in the
  proof assistant \Coq) and the metatheory.
\end{abstract}

\section{Introduction}
\label{sec:intro}

In a previous paper~\cite{lescanne06:_mechan_coq}, I have presented an
implementation of the \elck{} in \Coq{}.  There I have shown how
this applies to prove mechanically popular (and less popular) puzzles
as prolegomenon of other potential applications.  In these experiments
I have shown in particular that in the literature (mostly devoted to
study \emph{model theory} of \elck{}) some concepts of proof theory
are not clearly brought out and statements made at the meta-level,
i.e., in the meta-theory, are not sorted out from statements made at
the level of the language, i.e., in the theory.  In the deep embedding
in a proof assistant (where the logic is fully implemented into the
meta-language) the distinction between meta-theory and theory is made
explicit, by construction.  The proof assistant cannot accept
ill-formed expressions and forces the user to specify the level of
statements he makes, namely \emph{inside} the theory or \emph{outside}
the theory.  Thus the kind of implication or quantification or even
statement, e.g., axiom or premise of a logical implication, has to be made
precise.  On the opposite, in the handwritten treatments of the
puzzles, it is not clear whether a statement is made an axiom stated
as such in the meta-theory or a proposition stated as the premise of a
logical implication.  This confusion is especially present in the
literature on economic
games~\cite{samet96,geanakoplos94:_handb_game_theor}.  Using a
quantification in the meta-theory vs a quantification in the theory
can change dramatically the strength of a statement and its scope.

In this paper, my approach is this of a proof theorist with
inclination to experiments.  My goal is twofold.  First I present a
new axiomatization of \elck{} (axiom \FP{} and rule \LFP).  Second I
discuss a specific problem of \elck, namely the dilemma between
internalizing or externalizing implication.  Here one needs some
explanation.  In a proof theoretic approach there are two kinds of
implications: an internal implication (the implication of the object
theory) written here $?"=>"?$ , and the external implication (the
implication of the meta-theory) written $\frac{"|-"?}{"|-"?}$.  Here ,
$"|-"`v$ means \emph{``$`v$ is a theorem''}.  This discussion about
the two views of the same problem in common knowledge logic will be
made first through examples and at this exploratory state no
meta-theorem is proved.  There are two approaches when solving a
puzzle.  In the first approach, a statement is made an axiom, say
$"|-"`v$, this axiom leads to the proof of $"|-"`j$, proving the meta
implication $\frac{"|-"`v}{"|-"`j}$.  In the second approach, one
proves ${"|-" C_G(`v)"=>"`j}$, where $C_G$ is the \emph{common
  knowledge} modality.  From experiments, I have drawn the following
statements.  These two approaches seem to be equivalent and show the
interplay between the theory and the meta-theory.  An interesting
meta-theorem could be to prove that equivalence (see
Section~\ref{sec:equiv}).  I call \axvsck{} the equivalence of
$\frac{"|-"`v}{"|-"`j}$ with ${"|-" C_G(`v)"=>"`j}$.  In this paper
all the discussion is based on experiments made in the proof assistant
\Coq{} and the paper can be seen as the description of those
experiments.  I discovered in~\cite{FaginHMV95} that the
correspondence between $\frac{"|-"`v}{"|-"`j}$ and ${"|-" C_G(`v)"=>"
  C_G(`j)}$ is known, but it is not the one I am looking for.  In what
follows, the typewriter font is for code taken from the \Coq{}
implementation.  Most of the development in \Coq{} is available on the
WEB at
\url{http://perso.ens-lyon.fr/pierre.lescanne/COQ/epistemic_logic.v8}
(see ~\cite{lescanne06:_mechan_coq} or a presentation).
The rest can be found in~\cite{premaillon05:_logiq}.

\section{Presentation of \elck}
\label{sec:pres}

\subsection*{Historical facts}
The concept of common knowledge has been introduced by the philosopher
Lewis~\cite{lewis:1969} and since is used in several context namely
distributed systems~\cite{806736,moses86:_cheat}, artifical
intelligence~\cite{mccarthy77:_model_theor_knowl} and game
theory~\cite{aumann95}.

\subsection*{Epistemic logic}
\begin{figure}[t]
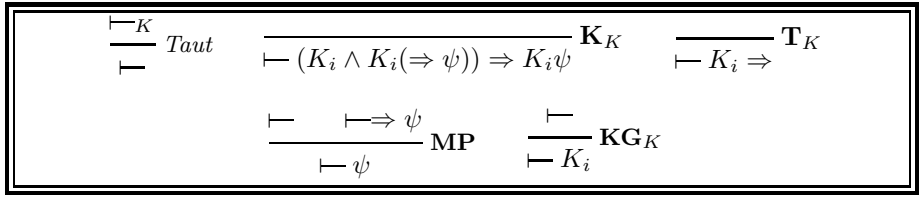

  \centering \doublebox{ \parbox{.70\textwidth}{\centering $
      \begin{array}{c@{\qquad}c@{\qquad}c}
        \prooftree
        "|-"_K `v %
        \justifies "|-" `v%
        \using \textit{\small Taut}
        \endprooftree
        &
        \prooftree
        \justifies "|-" (K_i `v \wedge K_i(`v "=>" `j)) "=>" K_i `j %
        \using \mathbf{K}_{K}
        \endprooftree 
        &
        \prooftree %
        \justifies "|-" K_i `v "=>" `v  %
        \using \mathbf{T}_K %
        \endprooftree
      \end{array}
      $

      \bigskip

      $
      \begin{array}{c@{\qquad}c}
        \prooftree
        "|-" `v \qquad  "|-" `v "=>" `j
        \justifies "|-" `j
        \using {\small \MP}
        \endprooftree 
        &
        \prooftree
        "|-" `v
        \justifies "|-"  K_i `v
        \using {\small \KG_K}
        \endprooftree 
      \end{array}
      $ }}
  \caption{The basic rules of epistemic logic: the system
    $\mathbb{T}$}
  \label{fig:basic-epi}
\end{figure}

The basis of \elck{} is \el.  In my experiments in
\Coq~\cite{BertotCasterant04}, \el{} is presented by a Hilbert-style
system of rules and axioms.  Since I use second order logic, I define
only the (internal) implication and I derive the other connectors.
There are only two rules namely \MP, i.e., the \emph{Modus Ponens} and
\KG{} also known as \emph{Knowledge Generalization} and three axioms
\emph{Taut}, \textbf{K} and \textbf{T}.  Actually \emph{Taut} is an
axiom scheme as it says that every classical tautology is a theorem in
\elck.  Such an approach requires a ``deep embedding'' (see
annex~\ref{sec:deepEmb}).  The main reason is that modal logic cannot
be easily implemented with natural deduction without changing its
basic philosophy (see annex~\ref{sec:boxing}).  Epistemic logic is
based on modal logic and in this paper only the system $\mathbb{T}$
(see Figure~\ref{fig:basic-epi}) is considered.  Since there is much
flexibility in the terminology, I decided to stick to the terminology
of~\cite{FaginHMV95}.  Epistemic logic introduces one modality for
each agent: it expresses that that agent ``knows'' the proposition
that follows the modality.  More specifically, if $`v$ is a
proposition, $K_i(`v)$ is the proposition $`v$ modified by the
modality $K_i$ which means \emph{``Agent $i$ knows~$`v$''}.  In
Figure~\ref{fig:basic-epi}, the statement $"|-"_K `v$ means that $`v$
is a theorem in classical propositional logic (this time, $K$ stands
for the German adjective ``klassisch''~\cite{gentzen35}).  Knowing
whether classical logic is relevant is a topics of research with René
Vestergaard.

\subsection*{\Elck}

Now let us suppose that we have a group $G$ of agents.  The knowledge
of a fact $`v$ can be shared by the group~$G$, i.~e., \emph{``each agent
  in $G$ knows $`v$''}.  We write $E_G(`v)$ and the meaning of $E_G$
is easily axiomatized by the equivalence given in
Figure~\ref{fig:shar} which can also be seen as the definition of
$E_G$; it is called \emph{shared knowledge}.

\begin{figure}[t]
  \centering \doublebox{ \parbox{.28\textwidth}{\centering $ \prooftree
      \justifies "|-" E_G(`v) "<=>" \bigwedge_{i`:G} K_i `v \using
      \textbf{E} \endprooftree $ }}
  \caption{Shared knowledge}
  \label{fig:shar}
\end{figure}

In \elck{}, there is another modality, called \textit{common
  knowledge} which is much stronger than shared knowledge.  It is also
associated with a group $G$ of agents and is written $C_G$.  Given
$`v$, $C_G(`v)$ is the least solution of the equation
\begin{eqnarray*}
  x &"<=>"& `v \wedge E_G(x).
\end{eqnarray*}
``Least'' should be taken w.r.t. the order induced by $"=>"$.  A
proposition $`j$ is \emph{less than} a proposition $`r$ if $`r"=>"`j$.
As well known in the fixed point theory, the least solution of the
above equation is also the least solution of the inequation:
\begin{eqnarray*}
  x &"=>"& `v \wedge E_G(x).
\end{eqnarray*}
The axiomatization of Figure~\ref{fig:com-know} characterizes
$C_G(`v)$ by two properties.  Together with the system $\mathbb{T}$
and the definition of $E_G$ it forms the system \ComK.  It asserts two
things.
\begin{enumerate}
\item $C_G(`v)$ is a solution of the inequation $x "=>" `v \wedge
  E_G(x)$, axiom \FP{},
\item If $`r$ is another solution of the inequation, then $`r$ implies
  $C_G(`v)$, which means that $`r$ is greater than $C_G(`v)$).  This is rule \LFP{}.
\end{enumerate}
One can prove that $C_G$ satisfies axioms and rules of $\mathbb{T}$,
where $K_i$ is replaced by $C_G$ even when $G=\emptyset$.  Thus we
prove
\[
\prooftree
\justifies "|-" (C_G `v \wedge C_G(`v "=>" `j)) "=>" C_G `j %
\using \mathbf{K}_C
\endprooftree 
\qquad
\prooftree %
\justifies "|-" C_G `v "=>" `v  %
\using \mathbf{T}_C %
\endprooftree
\qquad
 \prooftree
 "|-" `v
 \justifies "|-"  C_G `v \using \KG_C
 \endprooftree 
\]
                                 
$\KG_C$ stands for \emph{Common Knowledge Generalization}.  Notice
that $\mathbf{T}_C$ and $\frac{"|-"`v}{"|-"`v}$ on one side and $ "|-"
C_G `v "=>" C_G `v$ and $\KG_C$ on the other side form the two first
instances of \axvsck.  Actually one can prove more, namely that $C_G$
satisfies axiom $\four_C$, namely $"|-" C_G(`v)"=>" C_G(C_G(`v))$.
It~is a variant for \elck{} of the axiom $"|-" K_i(`v)"=>"
K_i(K_i(`v))$ of epistemic logic known as \emph{Positive
  Introspection} or~$\four_K$.  The proof of $\four_C$ does not
requires this of~$\four_K$\footnote{This seems to show that $\four$,
  which is a controverted axiom in general, should be stated more
  appropriately for the common knowledge of a group of agents rather
  than for the knowledge of an individual agent.}.

Notice that the presentation of common knowledge given in
Figure~\ref{fig:com-know} is new.  It is more robust than this of
Fagin et al.~\cite{FaginHMV95} which itself formalizes this of
Aumann~\cite{aumann95}.  Our axiomatization works even for an empty
set of agents and this is crucial, as starting with an empty set of
agents is the key of a recursive definition of $E_G$ and $C_G$;

\subsection*{Two presentations of \elck}

This presentation should be compared with this given by Meyer and van
der Hoek on page 46 of~\cite{meyer95:epist_logic} (see
Figure~\ref{fig:Mey-vdH}).  The system $\mathbb{T} \cup \{A7, A8, A9,
A10, R3\}$, together with the definition of $E_G$, is called \MvdH{}.
One notices that axioms $(A7)$ and $(A8)$ are just a splitting of
axiom \textit{Fixpoint}, i.e., one splits the conclusion $`v \wedge
E_G(C_G(`v))$.  Axiom $(A9)$ is  axiom $\textbf{K}_C$
mentioned above and $(R3)$ is $\KG_C$ also mentioned above.  As said, both
$(A9)$ and $(R3)$ can be proved as theorems in \ComK{}.  $(A10)$ is
more interesting and requires specific consideration.
Figure~\ref{fig:A10} sketches a proof of $(A10)$ as a theorem in
\ComK.  Therefore \ComK{} implies \MvdH{}.

\begin{figure}[t]
  \centering \doublebox{ \parbox{.33\textwidth}{\centering
      \begin{center}
        \begin{math}
          \prooftree \justifies "|-" C_G(`v) "=>" `v \wedge
          E_G(C_G(`v)) \using \FP{} \endprooftree
        \end{math}
      \end{center}

      \begin{center}
        \begin{math}
          \prooftree %
  "|-" `r "=>" `v \wedge E_G(`r)  %
  \justifies "|-" `r "=>" C_G(`v)
  \using \LFP{} %
  \endprooftree
\end{math}
\end{center}
}}
\caption{The rules for common knowledge}
\label{fig:com-know}
\end{figure}

\begin{figure}[t]
  \centering \doublebox{ \parbox{.44\textwidth}{
      \begin{center}
        \begin{math}
          \begin{array}{l@{\hspace*{5pt}}rcl}
            \mathit{(A7)} &   C_G(`v) & "=>" & `v\\[2pt]
            \mathit{(A8)} &   C_G(`v) & "=>" & E_G(C_G(`v))\\[2pt]
            \mathit{(A9)} &   C_G(`v)\wedge C_G(`v "=>" `j) & "=>" & C_G(`j)\\[2pt]
            \mathit{(A10)} & C_G(`v "=>" E_G(`v)) &"=>"& `v "=>" C_G(`v)\\[6pt]
            \mathit{(R3)} & \prooftree `v \justifies C_G(`v) \endprooftree
          \end{array}
        \end{math}
      \end{center}
    }}
  \caption{Meyer and van der Hoek axioms \MvdH}
  \label{fig:Mey-vdH}
\end{figure}

\begin{figure*}[t]
  \centering
  \doublebox{\parbox{.9\textwidth}{
      \begin{scriptsize}
        \begin{math}
          \prooftree%
      \prooftree
      \prooftree
  C_G(`v "=>" E_G(`v)) \wedge `v "=>" `v %
  \prooftree
  \prooftree
  C_G(`v "=>" E_G(`v))"=>" E_G(C_G(`v "=>" E_G (`v))) %
 \ \ovalbox{A8}  \quad
 \prooftree
 C_G(`v "=>" E_G(`v)) "=>" (`v "=>"  E_G(`v)) \ \ovalbox{A7}
  \justifies C_G(`v "=>" E_G(`v)) \wedge `v "=>"  E_G(`v)
  \endprooftree
  \justifies C_G(`v "=>" E_G(`v)) \wedge `v \ "=>"\ E_G (C_G(`v "=>" E_G (`v))) \wedge E_G(`v)
  \endprooftree
  \justifies C_G(`v "=>" E_G(`v)) \wedge `v \ "=>"\ E_G (C_G(`v "=>" E_G(`v)) \wedge `v) %
  \endprooftree 
  \justifies C_G(`v "=>" E_G(`v)) \wedge `v ~ "=>"~ `v \wedge E_G
  (C_G(`v "=>" E_G(`v)) \wedge `v) \using \textit{Transitivity of~} "=>"
  \endprooftree 
  \using \LFP{}
  \justifies C_G(`v "=>" E_G(`v)) \wedge `v "=>" C_G(`v)
  \endprooftree 
  \justifies  C_G(`v "=>" E_G(`v)) "=>" `v "=>" C_G(`v)
  \endprooftree 
\end{math}
\end{scriptsize}
}}
\caption{A proof of Meyer and van der Hoek's axiom $(A10)$}
\label{fig:A10}
\end{figure*}

\subsubsection*{\textit{\MvdH{} implies \ComK.}}

Indeed axiom \FP{} is an obvious consequence of \MvdH{}
and we prove that rule \LFP{} is a consequence of
\MvdH{} as follows.

\begin{displaymath}
  \prooftree 
  \prooftree 
  \prooftree 
  \prooftree 
  `r "=>" `v \wedge E_G(`r)  %
  \justifies `r "=>" E_G(`r)
  \endprooftree 
  \using (R3)
  \justifies  C_G(`r "=>" E_G(`r))
  \endprooftree 
  \using (A10 + \MP)
  \justifies `r "=>" C_G(`r)
  \endprooftree 
  \prooftree 
  \prooftree 
  \prooftree 
  `r "=>" `v \wedge E_G(`r)  %
  \justifies `r "=>" `v
  \endprooftree 
  \using (R3)
  \justifies C_G(`r "=>"`v))
  \endprooftree 
  \using (A9 + \MP)
  \justifies C_G(`r) "=>" C_G(`v)
  \endprooftree 
  \justifies `r "=>" C_G(`v) %
  \using (Transitivity of "=>")
  \endprooftree 
\end{displaymath}

\subsubsection*{\textit{$(R10)$ implies $(A10)$.}}

In the above proof, we should notice that instead of axiom $(A10)$, we
use  rule
\[
\begin{prooftree}
  C_G(`v "=>" E_G(`v)) \justifies `v "=>" C_G(`v) \using (R10)
\end{prooftree}
\]
which is a direct consequence of $(A10)$ by \MP.  By
analogy with $(A10)$, we call that rule $(R10)$. A closer look shows that
we use the derived rule
\[
\begin{prooftree}
  `v "=>" E_G(`v) \justifies `v "=>" C_G(`v) \using (R10')
\end{prooftree}
\]
which is the above rule combined with $(R3)$.  See section
\emph{Discussion} below to understand why we are interested in that
rule. Let us come back to $(R10)$ and let us call \MvdHalt{} the
system $\mathbb{T}\cup \{A7, A8, A9, R10, R3\}$.  Since we have a
proof of \ComK{} in \MvdHalt{} and a proof of \MvdH{}, in particular
of $(A10)$, in \ComK, we have an indirect proof of \MvdH{} in
\MvdHalt{} or, in short, of $(R10)$ implies $(A10)$.  Here is a direct
proof.

\newcommand{\abbrev}{A}

Let us state $\abbrev \ `=\ C_G(`v "=>" E_G(`v))$ in this proof.
First, let us prove $\abbrev \wedge `v "=>" C_G(\abbrev \wedge `v)$.
 
\begin{center}
\begin{scriptsize}
  \(
  \prooftree 
  \prooftree 
  \prooftree 
  C_G(`v "=>" E_G(`v)) "=>" E_G(C_G(`v "=>" E_G(`v))) \ \ovalbox{(A8)}%
  \justifies \abbrev \wedge `v "=>" E_G(\abbrev) %
  \endprooftree 
  \prooftree 
  \prooftree 
  C_G(`v "=>" E_G(`v)) "=>" (`v "=>" E_G(`v))\ \ovalbox{(A7)}
  \justifies  C_G(`v "=>" E_G(`v)) \wedge `v "=>" (`v "=>" E_G(`v)) \wedge `v  %
  \endprooftree 
  \quad %
  (`v "=>" E_G(`v)) \wedge `v \ "=>"\ E_G(`v) %
  \justifies C_G(`v "=>" E_G(`v)) \wedge `v "=>" E_G(`v) %
  \endprooftree 
  \justifies \abbrev \wedge `v \ "=>"\ E_G(\abbrev \wedge `v) %
  \endprooftree 
  \justifies \abbrev \wedge `v "=>" C_G(\abbrev \wedge `v) %
  \using (R10) %
  \endprooftree 
  \)
\end{scriptsize}
\end{center}

The rest is easy.  First, we notice that we have $C_G(\abbrev \wedge
`v) "=>" C_G(`v)$.

\[
\prooftree 
\prooftree 
\abbrev \wedge`v "=>" `v
\justifies C_G(\abbrev \wedge`v "=>" `v) \using (R3)
\endprooftree 
\justifies C_G(\abbrev \wedge`v) "=>" C_G(`v) \using (A9) + \MP
\endprooftree 
\]

By transitivity of $"=>"$, we get $\abbrev \wedge `v "=>" C_G(`v)$.
But clearly $\abbrev \wedge `v "=>" C_G(`v)$ is equivalent to $\abbrev
"=>" `v "=>" C_G(`v)$ which is $C_G(`v "=>" E_G(`v)) "=>" `v "=>"
C_G(`v)$, e.g., $(A10)$.

\subsection*{Discussion}
\label{sec:disc}

The equivalence between $(A 10)$ and $(R 10')$ is a third instance of
\axvsck.  Indeed, we have shown that a proposition of the form ${"|-"C_G(`r)
  "=>" `j}$ is equivalent to a rule of the form $\frac{"|-"
  `r}{"|-"`j}$.

\section{The three wise men}
\label{sec:1_exa}

The first example we address is the well-known example of the three
wise men.  See~\cite{lescanne06:_mechan_coq} for a more detailed presentation.
It is stated usually as follows (\cite{FaginHMV95},
Exercise~1.3): \emph{``There are three wise men.  It is common
  knowledge that there are three red hats and two white hats.  The
  king puts a hat on the head of each of the three wise men and asks
  them (sequentially) if they know the color of the hat on their head.
  The first wise man says that he does not know; the second wise man
  says that he does not know; then the third man says that he
  knows''}.  Let us call the three wise persons \textsf{Alice},
\textsf{Bob} and \textsf{Carol}.  Let us write \texttt{white Alice}
for \emph{``Alice wears a white hat''} and \texttt{red Alice} for
\emph{``Alice wears a red hat''}.  The puzzle is based on a function
which says whether an agent knows the color of her (his) hat:

\begin{verbatim}
Definition Kh := fun i => (K i (white i)) V (K i (red i)). 
\end{verbatim}
Clearly one has to prove that \texttt{Kh Carol} holds under some
assumptions.  To make clear theses assumptions, we define in addition
a few propositions namely
\begin{verbatim}
Definition One_hat := \-/(fun i:nat => white i | red i).
\end{verbatim}
which says that every agent wears a red hat or a white hat. If
\texttt{P} is a predicate, $\pourtout \mathtt{P}$ is the logical
quantification, i.e., the quantification in the theory not this in the
meta-theory.
\begin{verbatim}
Definition Two_white_hats := white Bob & white Carol ==> red Alice.
\end{verbatim}
which says that there are two white hats.  Notice that this is stated
in a weak form, indeed it is only when \textsf{Bob} and \textsf{Carol}
wear white hats that one can deduce that \textsf{Alice} wears a red
hat.  Moreover there are three concepts which say that each agent sees
the hat of the other agents and therefore knows the color of the hat.
\begin{verbatim}
Definition K_Alice_white_Bob :=  white Bob ==> K Alice (white Bob).
Definition K_Alice_white_Carol :=  white Carol ==> K Alice (white Carol).
Definition K_Bob_white_Carol :=  white Carol ==> K Bob (white Carol).
\end{verbatim}

\subsection*{A first result}
\label{sec:1res}
In a first attempt~\cite{lescanne06:_mechan_coq}, the five above
propositions were stated as axioms and I was able to prove:

\begin{alltt}
  |- K Carol (K Bob (\non Kh Alice) & \non Kh Bob)
              ==> K Carol (red Carol).
\end{alltt}

In \Coq{} this would give a statement like
\begin{alltt}
  |- One_hat & 
     K_Alice_white_Bob &
     K_Alice_white_Carol & 
     K_Bob_white_Carol &
     Two_white_hats  -> 
  |- K Carol (K Bob (\non Kh Alice) & \non Kh Bob) 
             ==> K Carol (red Carol).
\end{alltt}
where \texttt{->} is the meta-implication, i.e., this of \Coq{}
and as usual $\mathtt{|-}`v$ says that proposition~$`v$ is a
theorem.

\subsection*{A second result}
\label{sec:2res}

In the second attempt one proves:
\begin{alltt}
  |- K Carol (K Bob (One_hat &
                     K_Bob_white_Carol & 
                     K_Alice_white_Bob & 
                     K_Alice_white_Carol & 
                     (K Alice Two_white_hats) & 
                     \non Kh Alice) & 
              \non Kh Bob) 
    ==> Kh Carol.
\end{alltt}
This tells exactly the amount of knowledge which \textsf{Carol} requires to
deduce that she knows the color of her hat, actually red.  Let us
call \texttt{Alice\_Bob\_Carol} the group made of \textsf{Alice},
\textsf{Bob} and \textsf{Carol}. From the above statement, one derives
the corollary:
\medskip
\begin{alltt}
  |- C Alice_Bob_Carol (Two_white_hats & 
                        One_hat & 
                        K_Bob_white_Carol &
                        K_Alice_white_Bob & 
                        K_Alice_white_Carol) 
    ==> K Carol (K Bob (\non Kh Alice) & \non Kh Bob) ==> Kh Carol.
\end{alltt}
which is weaker.  But if we state
\begin{alltt}
  \begin{math}\varphi\equiv\end{math} Two_white_hats & 
      One_hat &
      K_Bob_white_Carol & 
      K_Alice_white_Bob &
      K_Alice_white_Carol
\end{alltt}
and 
\begin{alltt}
  \begin{math}\psi\equiv\end{math} K Carol (K Bob (\non Kh Alice) \& \non Kh Bob) ==> Kh Carol
\end{alltt}
we notice that we have exhibited a fourth instance of \axvsck{} since
${"|-"C_G(`v) "=>" `j}$ and $\frac{"|-" `v}{"|-"`j}$ are equivalent.

\section{The muddy children}
\label{sec:muddy}
This problem had many
variants~\cite{Littlewood1986,gardner84:_puzzl,gamow58:_puzzl,geanakoplos94:_handb_game_theor}.
It is a typical example of how a community of agents acquires
knowledge.  In its politically correct
version~\cite{FaginHMV95,meyer95:epist_logic}, a group of children
have mud on their head after playing during a birthday party.  The
kids do not know they have mud on their head.  The father of the kid
who organized the party asked the children to come around him in a
circle for the kids to see each other and he tells them that there is
at least one child who has mud on his face so that they clearly all
hear him.  Then Father asks the kids who have mud to step forward.  He
repeats this last sentence until all the kids step forward.

Philosophers have been puzzled by the fact that the first sentence of
Father namely \emph{``There is at least one child with mud on his
  face''} is absolutely necessary.  This fact is known by the
children, but by doing so, Father makes it a common knowledge.
In~\cite{lescanne06:_mechan_coq}, we have identified that the key
lemma is
\begin{alltt}
Lemma Progress :
    forall n p : nat,
    |- C ([:n+1:]) (At_least (n+1) p) & 
       E ([:n+1:]) (\non Exactly (n+1) p) 
       ==> C ([:n+1:]) (At_least (n+1) (p+1)).
\end{alltt}
In other words, if the fact that there is at least $p$ muddy children
is a common knowledge and all the children know that there is not
exactly $p$ muddy children, then  the fact that there is at least
$p+1$ muddy children is a common knowledge.  Together with the first
statement of Father:
\begin{alltt}
Axiom First_Father_Statement : 
    |- C ([:nb_children:]) (At_least n 1).
\end{alltt}
we are able to prove after $n$ steps \texttt{C ([:n:]) (At\_least n
  n)} which means that \emph{the fact that there is at least $n$ muddy
  children is common knowledge}.  This is the final result.  Common
knowledge is important here because one can ``progress'' in common
knowledge and not in shared knowledge.  Thus the first statement that
provides a first common knowledge allows initialization.  The proof of
\textsf{Progress} relies on a statement
\begin{alltt}
Knowledge_Diffusion :
 forall n p i : nat,
    |- E ([:n:]) (At_least n p) ==>
       E ([:n:]) (\non Exactly n p) ==> 
       K i (E ([:n:]) (\non Exactly n p)).
\end{alltt}
This statement is here to translate what children see after Father has
asked the muddy ones to step forward and none did.  They all know that
there is at least $p$ muddy children and they all know that there is
not exactly $p$ muddy children otherwise those with muddy face would
have stepped forward, but now each one knows that all the others know
that there is not exactly $p$ muddy children.

\subsection*{Knowledge\_Diffusion as an axiom}
\label{sec:muddy1}

In a first experiment, we made \texttt{Knowledge\_Diffusion} an axiom
and we were able to prove \texttt{Progress} in its above form.

\subsection*{Knowledge\_Diffusion as a common knowledge}
\label{sec:muddy2}

In the second experiment, we consider that proposition
\texttt{Knowledge\_Diffusion} should not be made an axiom, i.e., an
immutable principle, but it should be made just a rule of a game upon
everyone agrees.  Therefore the rules of the game are common knowledge
that everyone accepts; agreeing on these rules makes everyone to act and
reason according to them, i.e., ``rationally''.  In this version \emph{Progress} becomes:
\begin{alltt}
  Lemma Progress :
 forall n p : nat,
 |- C ([:n+1:])(Knowledge_Diffusion) ==> 
    (C ([:n+1:]) (At_least (n+1) p) & 
     E ([:n+1:]) (\non Exactly (n+1) p))
    ==> C ([:n+1:]) (At_least (n+1) (p+1)).
\end{alltt}

\subsection*{Discussion}
\label{sec:muddy-discussion}

Again we show that we can change an statement of the form $\frac{"|-"
  `v}{"|-"`j}$ into a statement of the form ${"|-"C_G(`v) "=>" `j}$.
Here
\begin{alltt}
   \begin{math}\varphi \equiv\end{math} C ([:n+1:]) (At_least (n+1) p) & 
        E ([:n+1:]) (\non Exactly (n+1) p))
\end{alltt}
and
\begin{alltt}
  \begin{math}\psi \equiv\end{math} C ([:n+1:]) (At_least (n+1) (p+1)).
\end{alltt}
This is a fifth instance of \axvsck.

\section{The equivalence between internal and external implication}
\label{sec:equiv}
  
Fagin et al~\cite{FaginHMV95} in exercise 3.29 notice, with no
reference, that $\frac{"|-" `v}{"|-"`j}$ and ${"|-"C_G(`v) "=>"
  C_G(`j)}$ are equivalent.  One notice by $\mathbf{T}_C$, i.e., $"|-"
C_G(`r) "=>" `r$, that this statement is stronger than \axvsck, which
states the equivalence between $\frac{"|-" `v}{"|-"`j}$ and
${"|-"C_G(`v) "=>" `j}$.  The proof of that result cannot be readily
implemented in \Coq{} in our current implementation of \elck{} since
this requires a deeper embedding of the theory.  In short, in order to
mechanize that proof, one needs not only internalize the object
implication, which we called internal implication, but also what we
called the external implication, since a meta-proof of the equivalence
requires an induction on the proof of $\frac{"|-" `v}{"|-"`j}$.  In a first step, one can
prove in \Coq{} that all the rules of \elck{}, namely \MP,
\KG{} and \LFP{} have their equivalent in the form
${"|-"C_G(`v) "=>" C_G(`j)}$, namely:

\[
"|-"C_G((`v "=>" `j) \wedge `v) "=>" C_G(`j) \qquad %
"|-" C_G(`v)"=>" C_G(K_i(`v)) %
\]
\[
"|-" C_G(`r "=>" `v \wedge E_G(`r))  "=>" C_G(`r "=>" C_G(`v))
\]
The first one is a variant, by the means of $"|-" C_G(`c\wedge`r)
"<=>" C_G(`c) \wedge C_G(`r)$, of $\mathbf{K}_C$ or $(A9)$.  The
second one is a basic result of \elck{}.  The third theorem has no equivalent in
the literature and has been proved in \Coq{} for that purpose.  Then
we get the following interesting result:
\newcommand{\CphiCpsi}{\ensuremath{\vdash C_G(\varphi) \Rightarrow
    C_G(\psi)}} %
\newcommand{\Cphipsi}{\ensuremath{\vdash C_G(\varphi)
    \Rightarrow \psi}}
\[
\xymatrix{
*++{\CphiCpsi}\ar[r] & *++++{\Cphipsi}\ar[r] & *++++{\frac{\vdash \varphi}{\vdash \psi}} \ar@(d,d)[ll]
}
\]

\bigskip

The back arrow is proved by induction of the length of the deduction
${"|-" `v~\texttt{->} "|-" `j}$.  Therefore, one notices three levels
of implications: the implication~$"=>"$ in the theory, the implication
$\frac{"|-"?}{"|-"?}$ in the metatheory and the implication
$\xymatrix{\ar[r]&}$ in the meta-metatheory.  From the above diagram one
gets
\[\xymatrix{*++++{\Cphipsi}\ar[r] & *++{\CphiCpsi}}.\]
Actually we have
\begin{displaymath}
  \prooftree  \Cphipsi \justifies  \CphiCpsi \endprooftree
\end{displaymath}
as follows
\begin{displaymath}
  \prooftree
\prooftree 
\Cphipsi \qquad "|-" C_G(`v) "=>" E_G(C_G(`v))
\justifies "|-" C_G(`v) "=>" `j \wedge E_G(C_G(`v))
\endprooftree
\justifies \CphiCpsi \using \LFP
\endprooftree
\end{displaymath}
since  $"|-" C_G(`v) "=>" E_G(C_G(`v))$ is a theorem of \elck{}.

\section{Conclusion}
\label{sec:concl}

On another hand, it is worth to mention the study on combining common
knowledge logic and dynamic logic we have done with Jérôme
Puisségur~\cite{puissegur05:_logiq,lescanne07_dynam_logic_of_common_knowl}.  The dynamic logic is used to
describe changes in the world, but those changes are \emph{purely
  epistemic} (an idea we borrow from Baltag, Moss and
Solecki~\cite{baltag98,baltag99}). This means that they affect only
knowledge of the agents and nothing else.  The muddy children puzzle
has been axiomatized in this framework and a proof of its results has
been fully mechanized in \Coq.  We can draw already two lessons form
those experiences.  First when merging two modal logics it seems that
internalizing common knowledge is more appropriate.  In other words,
an approach like ${"|-"C_G(`v) "=>" `j}$ should be preferred to
setting the axiom $"|-" `v$ to prove $"|-" `j$, as one does not know
which metatheory a specific statement belongs to: dynamic logic or
\elck?  Second a formalization of predicate logic, allows expressing
easily arbitrary depth of shared logic according to the number of
agents.  More precisely, common knowledge is not a priori necessary in
the muddy children example and just a specific number of imbricated
shared knowledge modalities corresponding to the number of children.
This fact was already noticed by
authors~\cite{geanakoplos94:_handb_game_theor}.

\subsubsection*{Acknowledgment}

I would like to thank Bertrand Prémaillon who made part of the
experiments in \Coq{} and René Vestergaard for stimulating
discussions.  


\appendix

\section{Deep embedding}
\label{sec:deepEmb}

A logic $\mathcal{L}$, the object logic or the object theory, is said
to be deeply embedded in another logic $\mathcal{M}$, the meta-theory,
or in a proof assistant if one considers the logic $\mathcal{M}$ to be
this of the proof assistant, if all the constituents of the logic
$\mathcal{L}$ are made objects of the logic $\mathcal{M}$ and all the
connectors and the rules of $\mathcal{L}$ are defined inside the logic
$\mathcal{M}$.  This is opposed to shallow embedding where
$\mathcal{L}$ and $\mathcal{M}$ may share connectors and rules.  A
shallow embedding is usually more concise, but in a deep embedding a
clear distinction is made between the connectors of the object theory
and those of the meta-theory.  In a deep embedding the connector and
the corresponding meta-connector can be somewhat connected, but
they cannot match completely.  For instance, it could happen that the
meta-disjunctions of two propositions meta-implies the proposition
made as the conjunction of the two propositions and not vice-versa, in
a sense made precise in formalizing the object theory.

Moreover not all the logics can be shallowly embedded.  This is the case
for \elck{} which cannot be formalized easily  in a natural deduction
framework (see next section).

\section{Why an Hilbert approach?}
\label{sec:boxing}

The reason why one cannot use a natural deduction of a sequent
calculus approach is essentially due to the rule \KG.  If one
accepts such a rule in natural deduction, one gets  
\[\prooftree `G"|-" `v \justifies K_i(`G) "|-" K_i(`v)\endprooftree\]
This requires to extend the operator $K_i$ to contexts like $`G$.  If
instead of $K_i$ one uses a modality $\Box$, one says that $\Box(`G)$
is a
\emph{``boxed context''}.  Actually \emph{linear logic}~\cite{LL}
is perhaps the archetypical modal logic and the equivalent of $K_i$ is
the modality \emph{of course} written ``!''.   The equivalent of
\KG{} is a rule called also \emph{of course}.  Without that rule
the proof net presentation is somewhat simple~\cite{lafont95from}.  Its introduction
requires a machinery of boxes which increases its complexity.

\end{document}